# Recognition of Grasp Points for Clothes Manipulation under unconstrained Conditions

Luz María Martínez and Javier Ruiz-del-Solar

Advanced Mining Technology Center & Dept. of Elect. Eng., Universidad de Chile
{luz.martinez, javier.ruizdelsolar}@amtc.cl

**Abstract.** In this work a system for recognizing grasp points in RGB-D images is proposed. This system is intended to be used by a domestic robot when deploying clothes lying at a random position on a table. By taking into consideration that the grasp points are usually near key parts of clothing, such as the waist of pants or the neck of a shirt. The proposed system attempts to detect these key parts first, using a local multivariate contour that adapts its shape accordingly. Then, the proposed system applies the Vessel Enhancement filter to identify wrinkles in the clothes, allowing to compute a roughness index for the clothes. Finally, by mixing (i) the key part contours and (ii) the roughness information obtained by the vessel filter, the system is able to recognize grasp points for unfolding a piece of clothing. The recognition system is validated using realistic RGB-D images of different cloth types.

## 1 Introduction

The development of service robotics has had an explosive increase during the last decade. In particular, new opportunities have been opened for the development of domestic robots, whose aim is to assist humans in daily, repetitive, or boring home tasks. One of these tasks is the handling of clothes, with the final goal of washing, ironing, folding, and storing the clothes according to the user's preference. When clothes are manipulated under unconstrained conditions, one of the key aspects to be solved is the determination of the grasp points to be used by the domestic robot.

In order to address this task, we propose a system for recognizing grasp points in RGB-D images, which is intended to be used by a domestic robot when deploying clothes lying at a random position on a table. By taking into consideration that the grasp points are usually placed near key clothes parts, such as the waist of pants or the neck in a shirt. The proposed system attempts to detect first these key regions using a local multivariate contour [3] that adapts its shape accordingly. Then, the proposed system applies the Vessel Enhancement filter [2] to identify wrinkles in clothing. Although this filter was originally designed to detect blood vessels on medical images based on their tubular geometric structure, in this work it is applied over the depth images in order to highlight the tubular shape of the wrinkles. Which allows us to compute a roughness index for the clothes. Then, by combining (i) the contours of the key parts and (ii) the roughness information obtained by the vessel filter, the system is able to recognize grasp points for unfolding a piece of clothing, which does not require having previously stored these points. We postulate that this approach obtains a better representation, than classical approaches based on the analysis of sliding windows or rectangular regions around a point of interest.

The proposed system can be of interest for researchers working in the development of similar systems (e.g. RoboCup @HOME and Amazon Picking Challenge teams), for their later integration in domestic service robots. The system is validated using realistic RGB-D images of different clothing types obtained from the Clothing Dataset [14] [16].

This paper is organized as follows. Section 2 presents some related work. Section 3 presents the proposed grasp detection system, and Section 4 shows the system evaluatation. Finally, discussion and conclusions are given in Section 5.

## 2   Related Work

There are many challenges associated with the manipulation of deformable objects, such as clothing. There are also different strategies to find the best way to grasp such objects. Some researchers prefer a two stage process, where the first task is to classify the object and then apply a pose recognition method [7] [8] [9].

Depending on the classification result, a proper garment pose recognition algorithm is applied. One approach is to use a *Dense* SIFT descriptor over depth images, obtaining invariance to rotation and scale, as an input to a *Basis Kernel function*(RBK) SVM on the first stage, and a *Linear Kernel function* SVM [7] [8] on the second. Another approach is to perform classification on both stages by means of a *deep Convolutional Neural Networks*(CNN) [9].

With the arrival of cheaper depth sensors, some reliable methods have been developed based on depth data. Particularly when dealing with a garment over a table; the following approaches have been proposed based on 3D descriptors: *Geodesic-Depth Histogram*(GDH) [13], *Fast Point Feature Histogram*(FPFH) [22] [23], *Heat Kernel Signature*(HKS) [15] and *Fast Integral Normal 3D*(FINDDD) [11]. However, once the piece of clothing hangs from a robot's gripper, it is common to take advantage of the classification and pose recognition procedure to identify optimal grasping points, which the robot can use to unfold the garment properly and efficiently.

Wrinkles are considered an important piece of information when working with clothes for the purpose of recognition, unfolding, or ironing. The "wrinkledness" measure has been widely used in state of the art algorithms, that use entropy measures to analyze how much of the surrounding area of a point has normals aligned in the same orientation, i.e. a flat surface or a combination of a few flat surfaces [12]. A more advanced analysis of wrinkles has also been carried out, aiming to identify their shape and topology using a size based classification procedure, which requires detecting the length, width and height of each wrinkle [21].

Trying to identify two grasping points for the robot to hold every garment class and ease its manipulation over a table, is an interesting problem. As a case of study, one approach developed to grasp a hand towel is to drag one hand along a towel border from a random point, until reaching a corner. Then, the other gripper repeats the process looking for an opposite corner [4] [5] [20] [6] [19]. Another approach is to train a *Hough Forest* for each garment class, where images are manually labeled with relevant grasping points [1].

## 3   Grasp Point detections system

Taking into consideration that the grasp points are usually near key clothing parts, such as the waist of pants or the neck in a shirt. We detect this key regions using a local multivariate contour [3] that adapts its shape accordingly, and the Vessel Enhancement filter [2] to identify wrinkles in clothing. This filter was designed to detect blood vessels on medical images, based on their tubular geometric structure; but when used in clothing depth images, it highlights the tubular shape of the wrinkles, allowing for the computation of a roughness index of the item. Combining the key part contours and the roughness information obtained by the vessel filter, we propose a system able to compute grasp points for unfolding a piece of clothing, which does not require having previously stored these points.

The proposed system is divided in three main parts: *Garment Key Part Recognition*, *Wrinkle Analysis* and the method that combines the previous steps called *Grasp Point Detections*. Figure 1 shows a general block diagram of the proposed system.

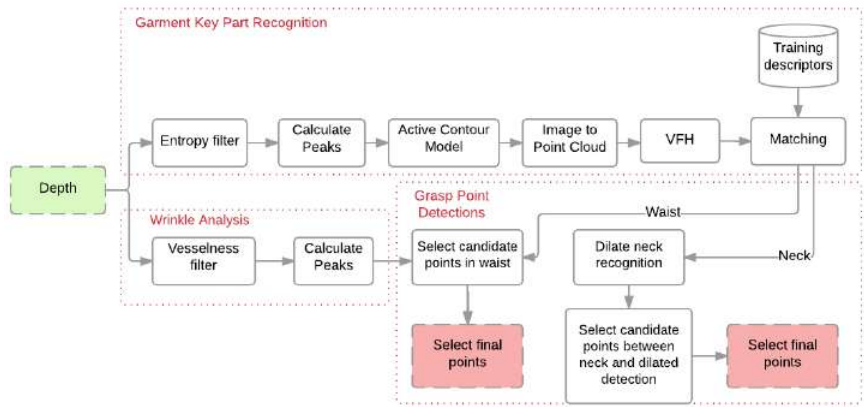

**Fig. 1.** Diagram of the grasp points detection system.

### 3.1 Garment key part recognition

The garment key part recognition it is based in VFH descriptors in a selected region performing a matching with k-nearest neighborhood. In this section a training phase is carried out by marking the region where the key part was observed on each image. Then a test phase where we search this key part in the image.

The VFH descriptor method [18] computes a global descriptor, which is formed by a histogram of the normal components of the object's surface. The histogram captures the shape of the object, and the viewpoint from which the point cloud is taken. First, the angles $\alpha$, $\phi$ and $\theta$ are computed for each point based on its normal and the normal of the point cloud's centroid $c_i$. The viewpoint-dependent component of the descriptor is a histogram of the angles between the vector $p_c - p_v$ and each normal point. The other component is a Fast Point Feature Histogram(FPFH) [17] estimated for the centroid of the point cloud, and an additional histogram of the distances of the points in the cloud to the cloud's centroid. The VFH descriptor is a compound histogram representing four different angular distributions of surface normals. In this work, the PCL implementation [10] is used, where each of those four histograms has 45 bins and the viewpoint-dependent component has 128 bins, totaling 308 bins.

In the recognition phase (Algorithm 1) the local maximums of the entropy filter image are considered as candidates points for contour evaluation in search of the key part of the clothing (see line 3).

**Algorithm 1:** Garment parts recognition

**1 Function** RecognitionGarmentPart(*img*)
   **Input :** Depth image to evaluate
   **Output:** Label of the garment part recognition
**2**   entropy = entropyFilter(*img*)
**3**   points = getPeaks(entropy)
**4**   listLabel = []
**5**   **for** *p in points* **do**
**6**     localRegion = getLocalRegion(*p, img*)
**7**     pcloud = *img2pointcloud*(*localRegion*)
**8**     normals = *getNormals*(*pcloud*)
**9**     vfh = *getVFHdescriptor*(*normals*)
**10**    neight, dist = knn.findNearest(vfh, 10)
**11**    listLabel.append( selectLabel(neight, dist))
**12**  **end**
**13**  **return** selectLabels(points, listLabel)

For contour evaluation, that is the selection of the contour of the part of the garment to be detected, we evaluated: edge detections and segmentation (Felenszwabs's, SLIC, Quickshift, Compact watershed, and Region Adjacency Graph). We select the active contours models [3], for its adaptation to the key part of the clothing (see line 6). This method consists of curves defined within an image domain that can move under the influence of internal forces coming from within the curve itself and external forces computed from the image data. The internal and external forces are defined so that the snake will conform to an object boundary or other desired features within an image. In Figure 2 we can see two examples of how the active contour models were adapted starting from a point that was in the key part.

In terms of the classification used for the recognition, we use k-NN with the 10 closest neighbors (see line 10), over VFH descriptors (see lines 6-9), where it was selected by a voting system. In the case where two or more classes had the same voting, the distances of the neighbors belonging to those classes were added and the one with the shortest distance was selected.

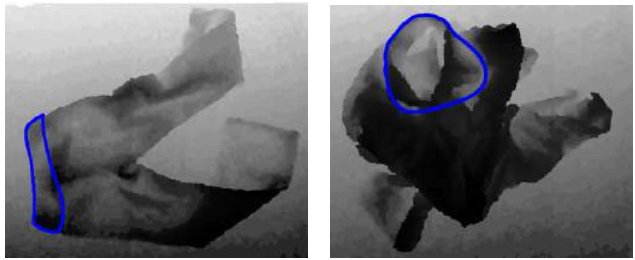

**Fig. 2.** Left: Countour in waist pant. Right: Countour in neck shirt.

### 3.2   Wrinkle analysis

The classical approach for the wrinkle analysis is based on an entropy filter, which measures how much of the surrounding area of the point has normals aligned in the same orientation i.e. a flat surface or a combination of a few flat surfaces [12].

The entropy measure is computed using a local descriptor based on the surface normals of a 3D point cloud. In particular, we use the inclination and azimuth angles in the spherical coordinates representation of the normal vectors:

$$(\phi, \theta) = \arccos\left(\frac{z}{r}\right), \arctan\left(\frac{y}{x}\right) \qquad (1)$$

$$r = \sqrt{x^2 + y^2 + z^2} \qquad (2)$$

Where $\phi$ is the inclination and $\theta$ is the azimuth, $(x, y, z)$ are the 3D point coordinates, and $r$ is the radius in spherical. Then, the distribution model is created in a bi-dimensional histogram with the inclination and azimuth values in a local region around each point(64 x 64 bins), and its entropy is computed as:

$$H(X) = -\sum_{i=1}^{n} p(x_i) \log p(x_i) \qquad (3)$$

This work proposes to exploit the geometric structure of clothes contours on depth images, by means of the Multiscale Vessel Enhancement Filter (MVEF) proposed by Frangi et al. This filter was originally designed to detect blood vessels on medical images, which have a tubular-like geometric structure [2].

The filter is applied to grayscale images and is based on a local *vesselness* measure, which is obtained by a second order approximation of the point and its neighborhood. The three lower Hessian eigenvalues are used to geometrically model the region by an ellipsoid, whose shape and orientation are defined by the eigenvalues and the related eigenvectors. Depending on the eigenvalue magnitudes, the filter can discriminate the region into plate, line and tubular-like structures. Furthermore, the author discusses the high noise and background suppression capabilities of the filter, in relation to the objective tubular structures.

In this work, depth images are used as the filter input, avoiding the complexity associated with highly textured cloth pieces, which can generate false positives. The filter highlights the tubular shape of the wrinkles and the folds generated by the clothes in the depth image. In order to account for patterns of different sizes, the filter strategy is to apply the described process at multiple scale levels, choosing the maximum local vesselness measure as the candidates points to grasp.

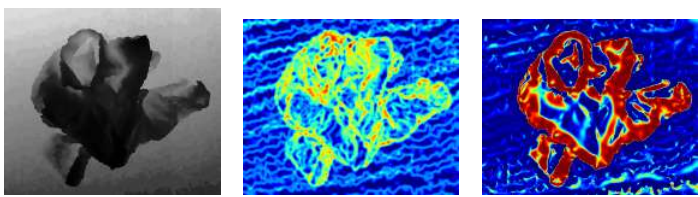

**Fig. 3.** Left: Input depth image. Center: Entropy filter. Right: Vessel filter

### 3.3   Grasp point detections

Taking the information from the detection of the key part and the local maxima on the wrinkles image, the grasp points are extracted in the following ways depending on the detected class:

- *T-Shirt and Shirt (Algorithm 2):* First, the local maximum points of the vessel filter image are obtained as candidate points (see line 2). Then, a morphological dilation is applied to the mask representing the neck and only those points that are within the dilated outline were filtered. To select between them, lines are created between all pairs of candidate points, selecting the one which passed closer to the contour center of the neck (see lines 4-13).

- *Pants (Algorithm 3):*  First, the local maximum points of the vessel filter image are obtained as candidate points (see line 2). Then, the pair of points of the waist detection contour that are further away are searched. To select the grasp points, we looked for the ones closer to the previous points while lying within the contour (see lines 4-13).

---

**Algorithm 2:** Select Grasp Points : T-Shirt and Shirt

**1 Function** selectPointsNeck(*mask, points*)
**2**     *center_point* = getCenterPoint(mask) ;
**3**     mask2 = dilate(mask);
**4**     filteredPoints = points in mask2 and not in mask ;
**5**     min_dist = -1 ;
**6**     **for** *p1 in filteredPoints* **do**
**7**         **for** *p2 in filteredPoints* **do**
**8**             *straight_line* = getStraightLine(*p1*, *p2*);
**9**             dist = distancePoint2Line(*center_point*, *straight_line*) ;
**10**            **if** *dist < min_dist or min_fist == -1* **then**
**11**                pointA, pointB = p1, p2 ;
**12**            **end**
**13**        **end**
**14**    **end**
**15**    **return** pointA, pointB;

---

**Algorithm 3:** Select Grasp Points : Pant

**Input   :** Mask of the garment part detection and the local maxima points of the wrinkle analysis with the vessel filter
**Output :** Two grasp points

**1 Function** selectPointsWaist()
**2**     *pe1*, *pe2* = getExtremePoint(mask)
**3**     min_dist = -1
**4**     **for** *p1 in filteredPoints* **do**
**5**         **for** *p2 in filteredPoints* **do**
**6**             *d1* = getEuclideanDist(*p1*, *pe1*) + getEuclideanDist(*p2*, *pe2*)
**7**             *d2* = getEuclideanDist(*p1*, *pe2*) + getEuclideanDist(*p2*, *pe1*)
**8**             *dist* = min(*d1*, *d2*)
**9**             **if** *dist < min_dist or min_fist == -1* **then**
**10**                pointA, pointB = p1, p2
**11**            **end**
**12**        **end**
**13**    **end**
**14**    **return** pointA, pointB;

---

## 4 Evaluation

### 4.1 Setup and Methodology

This work was evaluated using the Clothing Dataset [14] [16], which consists of depth pointcloud images of common deformable clothing pieces laying on a table.

Meaningful clothing parts are manually annotated with polygons. The data was recorded using a Kinect 3D camera, which provides 640x480 pixels color images and pointclouds in PCD v.7 format. The dataset also provides segmentation masks for foreground-background segmentation. We only considered 3 classes among the 6 available on the dataset: Pants, Shirt, T-Shirt.

Two parts of the proposed system were evaluated using the Clothing Dataset: garment key part recognition and grasp point detections. In the garment key part recognition process we used 100 images of each class, 300 in total; using 60% of them for training, and 40% for the testing phase. In the grasp point detections we used 40 images of each class, 120 in total, where the key parts were correctly identified and at least a single grasp point is visible.

### 4.2  Results

**Garment Parts recognition**

The results of the detection of the key parts of clothing can be observed in the confusion matrix in Table 1. In the table, a low detection rate is observed in the Neck T-Shirt class mainly because of the similarity it has with the Waist Pant class by its semicircular shape. Unlike Shirt class has a different shape and gets to get 74% recognition.

In the key parts recognition of all classes analyzed only the 24% is remained as no-detection. These detections are because a Voxel Grid filter is applied to the point cloud which reduces the number of points, but these points are not enough to continue with the algorithm. However in most cases works well, but in others the selected parameters eliminate too many points.

**Table 1.** Garment Parts recognition confusion matrix. Classes: Neck Shirt (NS), Neck T-Shirt (NTS), Waist Pant (W), No detection (ND).

|     | NS  | NTS | W   | ND  |
| --- | --- | --- | --- | --- |
| NS  | **74%** | 2%  | 0%  | 24% |
| NTS | 5%  | **29%** | 26% | 40% |
| W   | 5%  | 25% | **60%** | 10% |

**Grasp points detections**

In the implementation a rectangular region of length 51 was generated around the detected points and the selected points in the ground truth. Using the metric Intersection over Union(IoU) to measure the accuracy of the grasp points detections defined with the equation 4.

$$IoU = Intersection/Union \qquad (4)$$

Table 2 shows the average of the IoU metrics for the 120 images. Where all classes of the IoU average of best located point have a similar detection rate, but the T-Shirt class results get worse considering both points. This may mean that the selected images of the t-shirt class have the second point more occluded in proportion than the other classes.

**Table 2.** IoU results for the points grasp detections.

|                 | Pant | Shirt | T-Shirt |
|----------------:|:----:|:-----:|:-------:|
| Mean all points | 0.34 | 0.28  | 0.07    |
| Mean best point | 0.65 | 0.57  | 0.54    |

In the best point calculation is considering only the most accurate detected point of both points, as the average exceeds 0.5 of IoU, confirmed that at least one of the grasp points could serve to get unfold the clothes. In the particular results of each key part, the same proportions can be observed, ie the best detection results of grip points is in the Pant class, then in Shirt and finally T-Shirt.

However, only those detections with an IoU greater than 50% were considered correct. Obtaining the following results in the detection of 1 and 2 points.

**Table 3.** Recall of the points grasp detections.

|         | 1 point | 2 points |
|--------:|:-------:|:--------:|
| Shirt   | 55%     | 33%      |
| T-Shirt | 48%     | 29%      |
| Pant    | 63%     | 43%      |

In Table 3 it can be seen that at least one of the grasp points results close to or greater than 50% detection. This may be because in the selected images of the database, the key part of the clothing class and at least one of the grab points are always visible, but the second point may be hidden. In these cases, it is difficult for the system to detect a point near the ground truth.

In the Tables 2 and 3, the results of 1 point and best point detection are close enough, it can be considered that at least the 55% in the Shirt class, the 48% in the T-Shirt class and the 63% in the Pant class of the best point have a useful fit.

In the Figure 4 it is possible to observe the results of the detection of grasp points and the images of the different steps of the algorithm in images of different classes.

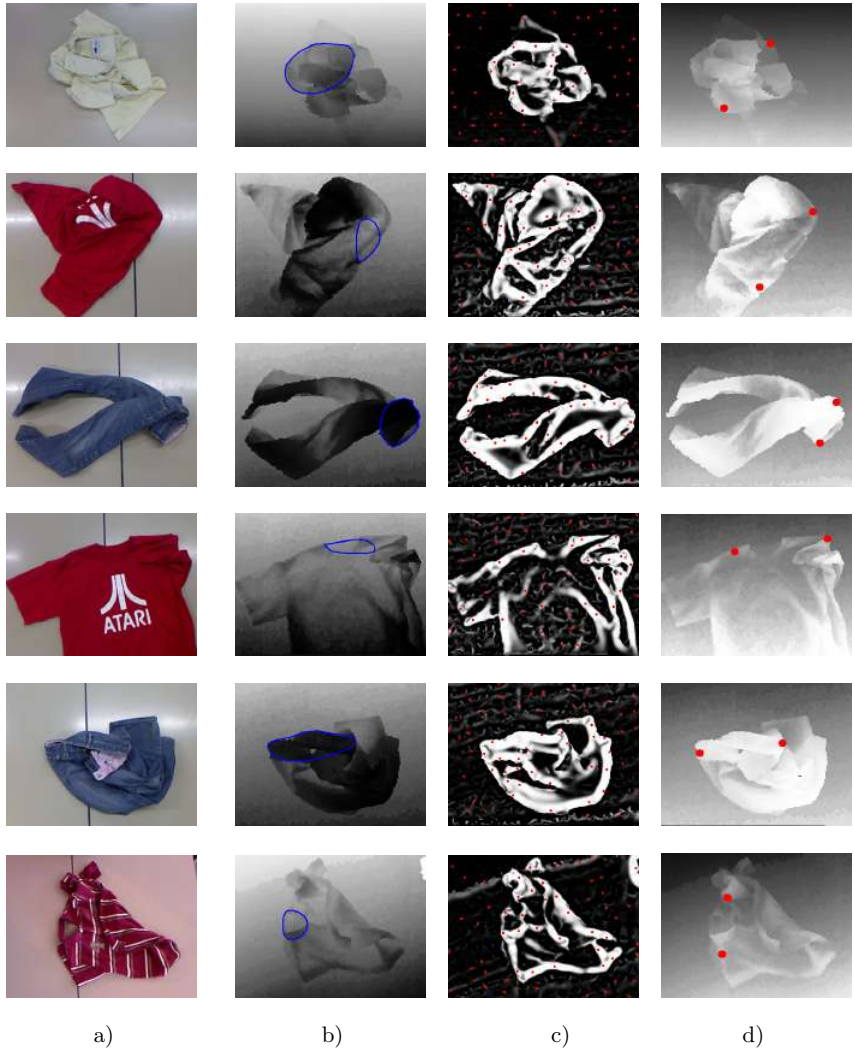

**Fig. 4.** Results of the algorithm. a) RGB image, b) depth image with the detected boundary of the key part marked in blue, c) wrinkle analysis with the vessel filter and the local maxima points in red, and d) detected grasp points in red.

## 5   Discussion and Conclusions

Grasp point detection is a complex problem because of its high dimensionality, given by all the positions that a piece of clothing can take when wrinkled. In this paper, we propose to exploit the geometry of wrinkles to solve this problem, based on the class of the detected part of clothing: for T-Shirt and Shirt classes the neck is searched, while the waist is considered for the Pant class.

The algorithm used the same procedure to detect points on the Shirt and T-shirt classes, which could be extrapolated to similar classes of clothes, such as the Sweater class. In addition, the method used for the detection of Pants points may be expanded for lower body garment classes, such as the Shorts class.

In relation to the grasp point detection results, at least one of the points has a high detection rate with respect to the points marked in the ground truth. This can be related to the selected database images, where the key part and at least one of the grasp points is always observed, while the second point can be occluded. However, it is observed that the detection rate of the T-Shirt class are lower compared to the other two classes. This can occur because of the difficulty of defining the neck contour in this class.

Another of the proposals of this paper is to use the resulting image of the vessel filter for wrinkle analysis, in order to calculate the "roughness" of a garment. There are two options proposed for calculating this metric, calculating the entropy index or calculating the average, both of which are calculated over the total area of the garment.

**Acknowledgments** This work was funded by CONICYT- PCHA/Doctorado Nacional/2014-21140280 and FONDECYT Project 1161500.